%% file: coling2016.tex
\documentclass[11pt]{article}
\usepackage{coling2016}
\usepackage{times}
\usepackage{url}
\usepackage{latexsym}

\def\svhline{%
  \noalign{\ifnum0=`}\fi\hrule \@height2\arrayrulewidth \futurelet
   \reserved@a\@xhline}

\usepackage{multicol}
\usepackage{abstract}
\usepackage{amsmath}
\usepackage{graphicx}
\usepackage{latexsym}
\usepackage{diagbox}
\usepackage{amssymb}

\usepackage{float}
\usepackage{multirow}
\usepackage{tabularx}

\usepackage{booktabs}
\usepackage{float}

\usepackage{amsmath} 

\usepackage{adjustbox}

\usepackage[table]{xcolor}
\usepackage{array}
\usepackage{multirow}
\usepackage{tabularx}
\usepackage{multicol}
\usepackage{multirow}
\usepackage{diagbox}
\usepackage{amsmath}
\usepackage{booktabs}
\usepackage{float}
\restylefloat{table}
\usepackage{enumitem}
\usepackage{makecell}
\usepackage{numprint}
\usepackage{siunitx}
\usepackage{hhline}

\title{Feature-Augmented Neural Networks for Patient Note De-identification}

\author{Ji Young Lee$^{1*}$, Franck Dernoncourt$^{1*}$, \"Ozlem Uzuner$^2$, Peter Szolovits$^1$ \\
  $^1$MIT, $^2$SUNY Albany\\
  {{\small  \tt \{jjylee,francky\}@mit.edu, ouzuner@albany.edu, psz@mit.edu}} \\
 {\small $^{*}$ These authors contributed equally to this work.}}
  \date{}

\begin{document}
\maketitle

\begin{abstract}
Patient notes contain a wealth of information of potentially great interest to medical investigators. However, to protect patients' privacy, Protected Health Information (PHI) must be removed from the patient notes before they can be legally released, a process known as patient note de-identification. The main objective for a de-identification system is to have the highest possible recall. Recently, the first neural-network-based de-identification system has been proposed, yielding state-of-the-art results. Unlike other systems, it does not rely on human-engineered features, which allows it to be quickly deployed, but does not leverage knowledge from human experts or from electronic health records (EHRs). In this work, we explore a method to incorporate human-engineered features as well as features derived from EHRs to a neural-network-based de-identification system. Our results show that the addition of features, especially the EHR-derived features, further improves the state-of-the-art in patient note de-identification, including for some of the most sensitive PHI types such as patient names. Since in a real-life setting patient notes typically come with EHRs, we recommend developers of de-identification systems to leverage the information EHRs contain.
\end{abstract}

\blfootnote{
     \hspace{-0.65cm} 
     This work is licenced under a Creative Commons 
     Attribution 4.0 International License.\\
     License details:
     \url{http://creativecommons.org/licenses/by/4.0/}
}

\section{Introduction and related work}
\label{intro}

Medical practitioners increasingly store patient data in Electronic Health Records (EHRs)~\cite{hsiao2011electronic}, which represents a considerable opportunity for medical investigators to construct novel models and experiments to improve patient care. Some governments even subsidize the adoption of EHRs, such as the Centers for Medicare \& Medicaid Services in the United States who have spent over \$30 billion in EHR incentive payments to hospitals and medical providers~\cite{ehrincentive}.

A legal prerequisite for a patient note to be shared with a medical investigator is that it must be de-identified. The objective of the de-identification process is to remove all Protected Health Information (PHI). Not appropriately removing PHI may result in financial penalties~\cite{desroches2013some,wright2013early}. In the United States, the Health
Insurance Portability and Accountability Act (HIPAA)~\cite{office2002standards} defines PHI types that must be removed, ranging from phone numbers to patient names. Failure to accurately de-identify a patient note would jeopardize the patient's privacy: the performance of a de-identification system is therefore critical.

A naive approach to de-identification is to manually identify PHI. However, this is costly~\cite{douglass2005identification,douglas2004computer} and unreliable~\cite{neamatullah2008automated}. Consequently, there has been much work developing automated de-identification systems. These systems are either based on rules or machine-learning models. Rule-based systems typically rely on patterns, expressed as regular expressions and gazetteers, defined and tuned by humans~\cite{berman2003concept,beckwith2006development,fielstein2004algorithmic,friedlin2008software,gupta2004evaluation,morrison2009repurposing,neamatullah2008automated,ruch2000medical,sweeney1996replacing,thomas2002successful}. 

Machine-learning-based systems train a classifier to label each
token as PHI or not PHI. Some systems are more fine-grained by detecting  which PHI type a token belongs to. Different statistical methods have been explored for patient note de-identification, including decision trees~\cite{szarvas2006multilingual}, log-linear models, support vector machines (SVMs)~\cite{guo2006identifying,uzuner2008identifier,hara2006applying}, and conditional random field (CRFs)~\cite{aberdeen2010mitre}. A thorough review of existing systems can be found in~\cite{meystre2010automatic,stubbs2015automated}. 

A more recent system has introduced the use of artificial neural networks (ANNs) for de-identification~\cite{dernoncourt2016identification}, and obtained state-of-the-art results. The system does not use any manually-curated features. Instead, it solely relies on character and token embeddings. While this allows the system to be developed and deployed faster, it fails to give users the possibility to add features engineered by human experts. 
Additionally, in practical settings of de-identification, patient notes typically come from a hospital EHR database, which contains metadata such as which patient each note pertains to, and other information such as the names of all doctors who work at the hospital where the patient was treated. The features derived from EHR databases may be useful for boosting the performance of de-identification systems.
In this work, we present a method to incorporate features to this ANN-based system, and show that it further improves the state-of-the-art.

\section{Method}
\label{sec:method}

The first model based on ANNs for patient note de-identification was introduced in~\cite{dernoncourt2016identification}: we extend upon their model. They utilized both token and character embeddings to learn effective features from data by fine-tuning the parameters.
The main components of the ANN model are Long Short Term Memories (LSTMs)~\cite{hochreiter1997long}, which are a type of recurrent neural networks (RNNs).

The model is composed of three layers: a character-enhanced token embedding layer, a label prediction layer, and a label sequence optimization layer. The character-enhanced token embedding layer maps each token into a vector representation. The sequence of vector representations corresponding to a sequence of tokens are input to the label prediction layer, which outputs the sequence of vectors containing the probability of each label for each corresponding token. Lastly, the sequence optimization layer outputs the most likely sequence of predicted labels based on the sequence of probability vectors from the previous layer. All layers are learned jointly. For more details on the basic ANN model, see~\cite{dernoncourt2016identification}.

\begin{figure*}[!ht]
\vspace{-0.1cm}
  \centering
      \includegraphics[width=0.7\textwidth]{{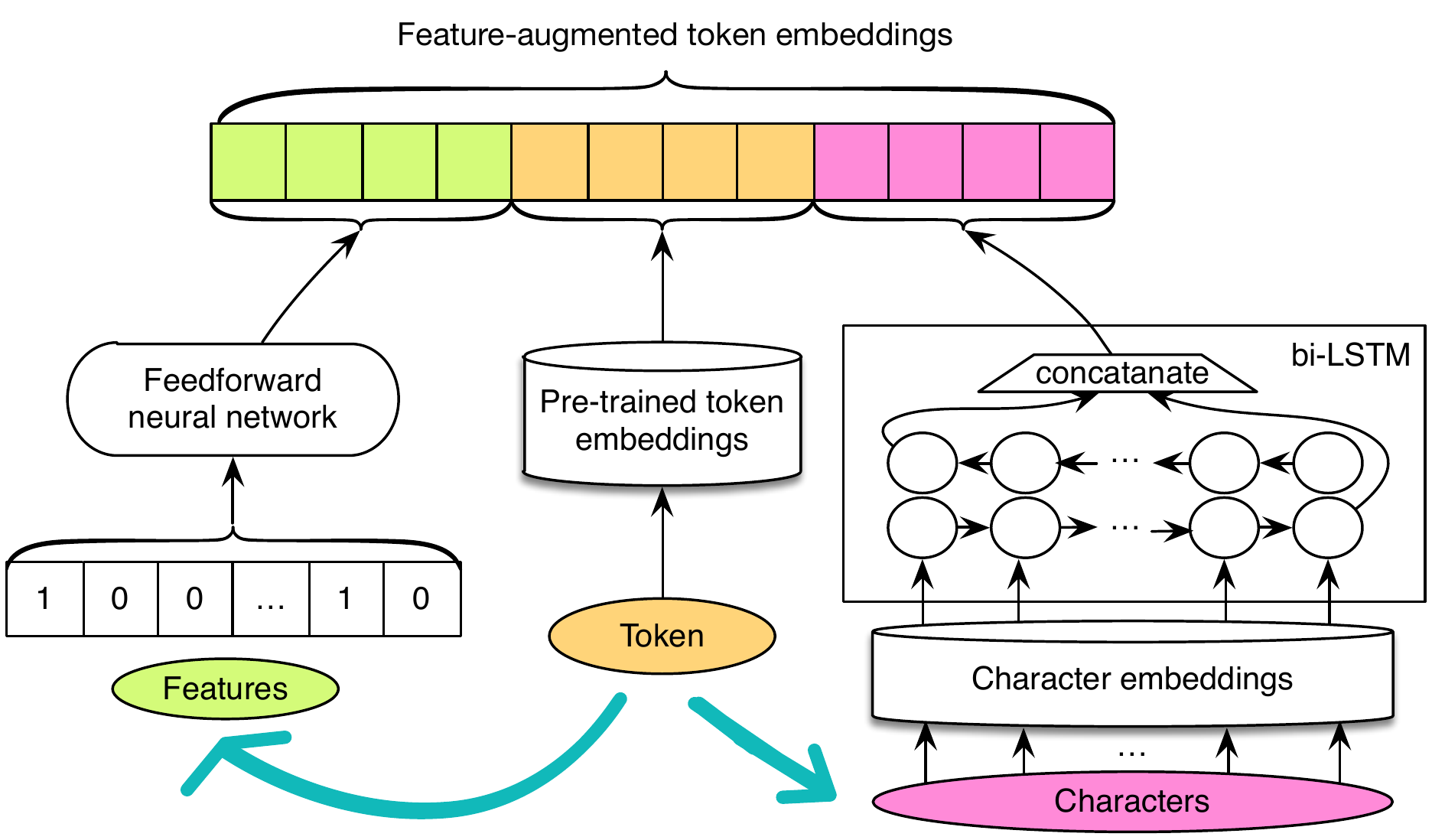}}
  \caption{Feature-augmented token embeddings. Each token is mapped to a token embedding that is the concatenation of three elements: the output of a feedforward neural network that takes the features as input, a pre-trained token embedding, and the output of a bidirectional-LSTM (bi-LSTM) that takes the character embeddings as input. }
  \label{fig:ann}
\end{figure*}

We augment this ANN model by adding features that are human-engineered or derived from 
EHR database, as presented in Table~\ref{tab:features}. The majority of human-engineered features are taken from~\cite{filannino2015temporal}, a few more features come from~\cite{yang2015automatic}, and additional gazetteers are collected using online resources. 
All features are binary and computed for each token. The binary feature vector comprising all features for a given token is fed into a feedforward neural network, the output vector of which is concatenated to the corresponding token embeddings, at the output of the character-enhanced token embedding layer, as Figure~\ref{fig:ann} illustrates.

\input{tables/features}

\section{Experiments}
\label{sec:experiments}

We evaluate our model on the de-identification dataset introduced in~\cite{dernoncourt2016identification}, which is a subset of the MIMIC-III dataset~\cite{goldberger2000physiobank,saeed2011multiparameter,johnson2016mimic}, using the same train/validation/test split (70\%/10\%/20\%). We chose this dataset as each note comes with metadata, such as the patient's name, and it is the largest de-identification dataset available to us. 
It contains 1,635 discharge summaries, 2,945,228
tokens, 69,525 unique tokens, and 78,633 PHI tokens. 

The model is trained using stochastic gradient descent, updating all parameters, i.e., token embeddings, character embeddings, parameters of bidirectional LSTMs, and transition probabilities, at each gradient step. For regularization, dropout is applied to the character-enhanced token embeddings before the label prediction layer. We set the character embedding dimension to 25, the character-based token embedding LSTM dimension to 25, the token embedding dimension  to 100, the label prediction LSTM dimension to 100, the dropout probability to 0.5, and we use GloVe embeddings~\cite{pennington2014glove} trained on Wikipedia and Gigaword 5~\cite{parker2011english} articles as pre-trained token embeddings. The hyperparameters were optimized based on the performance on the validation set.

\section{Results}
\label{sec:results}

Table~\ref{tab:results} presents the main results. The epochs for which the results are reported are optimized based on either the highest F1-score or the highest recall on the validation set. As expected, choosing the epoch based on the recall improves the recall on the test set, while lowering the precision. Overall, adding features consistently improves the results.

\input{tables/results}

\input{tables/results-details}

Table~\ref{tab:results-detailed} details the results for each PHI type. The system using only the EHR features yields the highest recall for 6 out of 12 PHI types. Most importantly, the recall for patient and doctor names are higher when using features than when using no feature: this is expected as the patient name of the note and the doctor names are used as features. In fact, the two remaining false negatives for patient names are annotation errors. For example, in the sentence ``The patient responded well to \emph{Natrecor} in the past, but the improvement disappeared soon'', the drug name \emph{Natrecor} was incorrectly marked as a patient name by the human annotator. This result is highly remarkable as patient names are the most sensitive information in a patient note~\cite{south2014evaluating}.

Adding all features often lowers the recall compared to using EHR features only, although the F1-score remains virtually unchanged. 
This is somewhat surprising, as we had expected that the features would help, as using the same feature set with a CRF to perform de-identification yields state-of-the-art results next to the ANN models~\cite{dernoncourt2016identification}.  
This could be explained as follows. Human-engineered features tend to have higher precision than recall, as it is often hard to design regular expressions or gazetteers that can detect all possible instances or variations of the desired entities. 
We conjecture that as the ANN model learn to rely more on such features, it might lose the ability to learn to pick up tokens that deviate from engineered features, resulting in a lower recall. 
For example, we notice that the phone PHI tokens that are not detected by the model using all features but are detected by the other two models, are ill-formed phone numbers such as ``617-554-$\mid$2395'', or phone extensions such as ``617-690-4031 ext 6599''. 
Since the phone regular expressions do not capture these two examples, they are more likely to be false negatives in the model that uses the phone regular expression features.

\vspace{-0.2cm}
\section{Conclusion}
\label{sec:conclusion}
\vspace{-0.2cm}
In this paper we presented an extension of the ANN-based model for patient note de-identification that can incorporate features. We showed that adding features results in an increase of the recall, in particular features leveraging information from the associated EHRs, namely patient names and doctor names.

Our results suggest that constructing patient note de-identification systems should be performed using structured information from the EHRs, the latter being available in a typical, real-life setting. We restricted our EHR-derived features to patient and doctor names, but it could be extended to the many other structured fields that EHR contain, such as patients' addresses, phone numbers, email addresses, professions, and ages.

\section*{Acknowledgements}

The project was supported by Philips Research. The content is solely the responsibility of the authors and does not necessarily represent the official views of Philips Research. We warmly thank Michele Filannino, Alistair Johnson, Li-wei Lehman, Roger Mark, and Tom Pollard for their helpful suggestions and technical assistance.

\bibliographystyle{acl}
\bibliography{coling2016}

\end{document}

%% file: tables/features.tex
\begin{table} [ht]
\footnotesize
\centering
\setlength{\extrarowheight}{3pt}
\setlength{\arraycolsep}{5pt}
\newcolumntype{L}{>{\arraybackslash}p{12cm}}
\begin{tabular}{|l|L|}
\hline
\textbf{Feature types} & \textbf{Features} 	\\
\hline
\hspace{-0.2cm}\begin{tabular}{l}
\text{Note metadata} \\ 
\text{Hospital data}
\end{tabular} &
\hspace{-0.1cm}$\left.\begin{tabular}{@{\ }l@{}}
    \text{Patient's first name, patient's last name} \\ \text{Doctor's first names, doctor's last names}
  \end{tabular}\right\}$ EHR features			\\
\hline
\text{Morphological}					& Ends with s, is the first letter capitalized, contains a digit, is numeric, is alphabetic, is alphanumeric, is title case, is all lower case, is all upper case, is a stop word 		 \\
\text{Semantic/Wordnet}	& Hypernyms, senses, lemma names				\\
\text{Temporal}				& Seasons, months, weekdays, times of the day, years, years followed by apostrophe, festivity dates, holidays, cardinal numbers, decades, fuzzy quantifier (e.g., ``approximately'', ``few''), future trigger (e.g., ``next'', ``tomorrow'') 	\\	
\text{Gazetteers}	& Honorifics for doctors, honorifics, medical specialists, medical specialties, first names, last names, last name prefixes, street suffixes, US cities, US states (including abbreviations), countries, nationalities, organizations, professions			\\
\text{Regular expressions}	& \text{Email, age, date, phone, zip code, id number, medical record number}				\\
\hline
\end{tabular}
\caption{Feature list. Note metadata and hospital data are derived from the EHR database. Morphological, semantic/wordnet, and temporal features are commonly used features for NLP tasks. Gazetteers and regular expressions are specifically engineered for the task. }\label{tab:features}
\end{table}

%% file: tables/results.tex
\newcolumntype{?}{!{\vrule width 1pt}}
\newcolumntype{C}{>{\centering\arraybackslash}X}%
\newcolumntype{M}{N{2}{2}}
\renewcommand*\arraystretch{1.1}
\begin{table}[h]
\vspace{-0.2cm}
\caption{Binary HIPAA token-based results (\%) for the ANN model, averaged over 5 runs. The metric refers to the detection of PHI tokens versus non-PHI tokens, amongst PHI types that are defined by HIPAA.  ``No feature'' is the model utilizing only character and word embeddings, without any feature. ``EHR features'' uses only 4 features derived from EHR database: patient first name, patient last name, doctor first name, and doctor last name. ``All features'' makes use of all features, including the EHR features as well as other engineered features listed in Table~\ref{tab:features}. ``Optimized by F1-score'' and ``optimized by recall'' means that the epochs for which the results are reported are optimized based on the highest F1-score or the highest recall on the validation set, respectively.}
\label{tab:results}
\npdecimalsign{.}
\nprounddigits{2}
\begin{tabularx}{\textwidth}{|l|CCC|CCC|}
\hline
\multirow{2}{*}{  } & \multicolumn{3}{c|}{Binary HIPAA (optimized by F1-score)} & \multicolumn{3}{c|}{Binary HIPAA (optimized by recall)} \\
  & {Precision} & {Recall} & {F1-score} & {Precision} & {Recall} & {F1-score} \\
 
\hline
No feature 		& 99.103 & 99.197& 99.150 & 98.557 & 99.376 & 98.965 \\
EHR features 	& 99.100 & 99.304& 99.202 & 98.771 & \textbf{99.441} & 99.105 \\
All features 	& \textbf{99.213} & \textbf{99.306}& \textbf{99.259} & \textbf{98.880} & 99.420 & \textbf{99.149} \\

\hline
\end{tabularx} 
\npnoround
\end{table}

%% file: tables/results-details.tex
\newcolumntype{?}{!{\vrule width 1pt}}
\newcolumntype{C}{>{\centering\arraybackslash}X}%
\renewcommand*\arraystretch{1.1}
\begin{table}[h]
\caption{\vspace{-0.0cm}Binary token-based results (\%)
. The reported results are optimized by recall, and averaged over 5 runs. The symbol \textsuperscript{1} indicates that the PHI type is not required by HIPAA. The PHI type ``location'' designates any location that is not a street name, zip code, state or country. P stands for precision, R for recall, and F1 for F1-score.\vspace{-0.2cm} }
\vspace{-0.2cm}
\label{tab:results-detailed}
\centering
\npdecimalsign{.}
\nprounddigits{2}
\begin{tabularx}{0.95\textwidth}{|l|MMM|MMM|MMM|r|}
\hline
\multirow{2}{*}{  } & \multicolumn{3}{c|}{No feature} & \multicolumn{3}{c|}{EHR features} & \multicolumn{3}{c|}{All features} &  \\
 
 & \multicolumn{1}{c}{P} & \multicolumn{1}{c}{R} & \multicolumn{1}{c|}{F1} & \multicolumn{1}{c}{P} & \multicolumn{1}{c}{R} & \multicolumn{1}{c|}{F1} & \multicolumn{1}{c}{P} & \multicolumn{1}{c}{R} & \multicolumn{1}{c|}{F1} & {Support} \\
\hline
Zip & {100.0}	& {100.0}	& {100.0}	& {100.0}	& {100.0}	& {100.0}	& {100.0}	& {100.0}	& {100.0}	& 24\\
Date & 98.895	& 99.771	& 99.331	& 98.947	& {\fontseries{b}\selectfont} 99.785	&{\fontseries{b}\selectfont} 99.364	& {\fontseries{b}\selectfont}98.989	& 99.691	& 99.339	& 20627\\
Phone& 98.308	& {\fontseries{b}\selectfont} 99.583	& 98.939	& {\fontseries{b}\selectfont}98.980	& 99.458	& 99.217		& 99.417	& 99.319	& {\fontseries{b}\selectfont}99.367	& 1438\\
Patient& 96.893	& 98.344	& 97.606	& 98.617	& 99.139	& 98.877	& {\fontseries{b}\selectfont}99.207	& {\fontseries{b}\selectfont} 99.272	& {\fontseries{b}\selectfont}99.239	& 302\\
ID& 99.569	& 98.235	& 98.898	& 99.310	& {\fontseries{b}\selectfont} 98.824	& {\fontseries{b}\selectfont}99.066& {\fontseries{b}\selectfont}99.767	& 97.974	& 98.862	& 612	\\
Doctor\textsuperscript{1} & 97.467	& 98.172	& 97.818	& 97.265	& {\fontseries{b}\selectfont} 98.482	& 97.870	& {\fontseries{b}\selectfont}97.555	& 98.199	& {\fontseries{b}\selectfont}97.875	& 3676\\
Location& 96.018	& 95.714	& 95.857	& 96.413	& {\fontseries{b}\selectfont} 96.494	& 96.452	& {\fontseries{b}\selectfont}96.652	& 96.320	& {\fontseries{b}\selectfont}96.461	& 462\\
Age $\geq$ 90		& 75.120	& 94.286	&83.597	& 77.043	& {\fontseries{b}\selectfont} 95.715	& {\fontseries{b}\selectfont}85.353	 & {\fontseries{b}\selectfont}78.929	& 93.572	& 84.802 & 28\\ 
Hospital\textsuperscript{1}& 94.778	& 95.393	& 95.075& 94.771	& {\fontseries{b}\selectfont} 95.520	& 95.139	& {\fontseries{b}\selectfont}95.532	& 95.504	& {\fontseries{b}\selectfont}95.506	& 1259\\
State\textsuperscript{1}& 99.360	& {\fontseries{b}\selectfont} 94.328	& {\fontseries{b}\selectfont}96.759	& {\fontseries{b}\selectfont}99.683	& 94.030	& 96.731	& 99.394	& 91.940	& 95.491	& 67\\
Street & 96.774	& 85.246	& 90.539	& {\fontseries{b}\selectfont}97.627	& 85.246	& {\fontseries{b}\selectfont}90.961	& 93.908	& {\fontseries{b}\selectfont} 86.557	& 89.811	& 61\\
Country\textsuperscript{1} & 87.513	& 85.000	& 86.111	& {\fontseries{b}\selectfont}89.286	& 82.500	& 85.667		& 86.869	& {\fontseries{b}\selectfont} 95.000	& {\fontseries{b}\selectfont}90.560	& 16 \\
\hline
Binary & 98.407 & 99.192 & 98.798 & 98.478 & {\fontseries{b}\selectfont} 99.274 & 98.874 & {\fontseries{b}\selectfont}98.609 & 99.146 & {\fontseries{b}\selectfont}98.876 & 28572 \\

\hline
\end{tabularx}
\npnoround
\end{table}